\documentclass{article}

\usepackage{PRIMEarxiv}

\usepackage[utf8]{inputenc} 
\usepackage[T1]{fontenc}    
\usepackage{hyperref}       
\usepackage{url}            
\usepackage{booktabs}       
\usepackage{amsfonts}       
\usepackage{nicefrac}       
\usepackage{microtype}      
\usepackage{lipsum}
\usepackage{fancyhdr}       
\usepackage{graphicx}       
\usepackage{subcaption}      
\usepackage[backend=biber,doi=true,style=numeric,sorting=none,url=false]{biblatex} 
\graphicspath{{media/}}     

\title{Multimodal video analysis for crowd anomaly detection using open access tourism cameras}

\author{
  Alejandro Dionis-Ros, Joan Vila-Franc{\'e}s, Rafael Magdalena-Benedicto,   \\
  \textbf{Fernando Mateo, Antonio J. Serrano-L{\'o}pez}\\
  Intelligent Data Analysis Laboratory (IDAL) \\
  University of Valencia \\
  Valencia, Spain \\
  \texttt{\{alejandro.dionis, joan.vila, rafael.magdalena,} \\
  \texttt{fernando.mateo, antonio.j.serrano\}@uv.es} \\
}

\begin{document}
\maketitle

\begin{abstract}
    In this article, we propose the detection of crowd anomalies through the extraction of information in the form of time series from video format using a multimodal approach. Through pattern recognition algorithms and segmentation, informative measures of the number of people and image occupancy are extracted at regular intervals, which are then analyzed to obtain trends and anomalous behaviors. Specifically, through temporal decomposition and residual analysis, intervals or specific situations of unusual behaviors are identified, which can be used in decision-making and improvement of actions in sectors related to human movement such as tourism or security.

    The application of this methodology on the webcam of \textit{Turisme Comunitat Valenciana} in the town of Morella (\textit{Comunitat Valenciana}, Spain) has provided excellent results. It is shown to correctly detect specific anomalous situations and unusual overall increases during the previous weekend and during the festivities in October 2023. These results have been obtained while preserving the confidentiality of individuals at all times by using measures that maximize anonymity, without trajectory recording or person recognition.
\end{abstract}


\section{Introduction}
    In today's digital era, the amount of generated and stored data is growing at an exponential rate, with unstructured data types standing out among them. These data types, including text, images, audio, and video, offer new sources of knowledge. Applying a multimodal analysis approach, as applied in \cite{ding_mst-gat_2023, pmlr-v95-guo18a, nedelkoski_anomaly_2019}, involves combining different data formats and thus allows leveraging the richness of information inherent in each format, providing a more comprehensive and accurate understanding of the phenomena under study.

    Due to the high complexity of unstructured data, exploring the application of anomaly detection techniques becomes essential. These techniques allow identifying unusual patterns or atypical behaviors within datasets that can be extremely heterogeneous and difficult to interpret using conventional methods.

    In \cite{Shaukat_2021}, a literature review is conducted, and a taxonomy for anomaly detection in time series is proposed. This taxonomy distinguishes according to the type of anomaly to be detected, which can be point anomalies, contextual anomalies, or collective anomalies; the characteristics of the series, i.e., whether dealing with univariate or multivariate series; the context of the data, i.e., considering the spatio-temporal characteristics of the series in the analysis; and the methodology used, either through machine learning or statistical methods. Additionally, it also provides a review of the challenges that may be encountered when using these techniques, such as defining the normality of the data, the underrepresentation of anomalies in datasets, or the need to differentiate anomalies from possible noise present in the series.

    In \cite{ren_time-series_2019}, a system based on the Spectral Residual (SR) algorithm and Convolutional Neural Networks (CNN) is proposed, allowing online anomaly detection in business metric time series. In \cite{laptev_generic_2015}, a system for real-time anomaly detection is proposed, capable of detecting different types of anomalies due to the integration of various anomaly detection and prediction models. In \cite{geiger_tadgan_2020}, a model for unsupervised anomaly detection based on Generative Adversarial Networks (GANs) is proposed, with both the actors and critics of these networks based on LSTM recurrent networks, where series reconstruction allows anomaly detection. In \cite{xu_anomaly_2022}, a model based on Transformers for unsupervised anomaly detection is proposed. In this case, a version of Transformer with modified attention mechanism (Anomaly-Attention) is implemented, aiming to overcome the limitations of these in the task of anomaly detection. 
    
    In \cite{xue_real-time_2020}, a methodology for real-time anomaly detection in video is introduced. This approach utilizes feature maps, extracted from the energy of each frame, enabling the detection of object movement speed. Consequently, frames where the energy undergoes abrupt changes are identified as anomalous. In \cite{nazir_suspicious_2023}, a methodology for detecting suspicious behaviors in video is presented. Through object detection and tracking algorithms, the spatio-temporal characteristics of individuals in the video are extracted. Subsequently, the series obtained from these characteristics are classified based on whether they exhibit anomalous behaviors or not.

    In this article, the application of statistical techniques for anomaly detection in time series is proposed, using a multimodal approach. These series, obtained from video format data, represent the number of people detected in a given time interval and the percentage of heatmap saturation of image occupancy. Considering as anomalies those intervals where the number of detected people or the saturation percentage have values far from the rest of the intervals, we obtain as a result the intervals where these unusual behaviors have been detected, which can be used in decision-making and process improvement in sectors such as tourism or video surveillance. By foregoing tracking algorithms and applying measures that preserve anonymity during the analysis, the confidentiality of individuals is maintained at all times.

\section{Dataset}
    \subsection{Source of the data}
        The Valencian Tourism Agency has deployed over 50 web cameras in the \textit{Comunitat Valenciana} since 2001, in a project sponsored by Tourism. These cameras are accessible on the tourist portal \textit{Webcams de la Comunitat Valenciana\footnote{\href{https://www.comunitatvalenciana.com/es/webcams}{https://www.comunitatvalenciana.com/es/webcams}}} and broadcast live throughout the day images of various destinations, offering users the opportunity to view the status of these places in real time. Strategically located in collaboration with municipalities and sector entrepreneurs, they allow users to follow events, festivities, and sports competitions.

        Specifically, the data used has been obtained from the town of Morella\footnote{\href{https://en.wikipedia.org/wiki/Morella,_Castellón}{https://es.wikipedia.org/wiki/Morella\_Castell\'on}}, Spain. The geographical location of Morella has been crucial throughout the centuries and historical events, which is why it is currently considered a population of high tourist value. The webcam located in this town consists of a static camera whose broadcast is done at a resolution of 1920x1080 pixels, at a rate of 30 frames per second.
        
        Currently, said camera has been upgraded and features different characteristics from the previous one. Because of this, the amount of data available for the study is limited (from 20/09/2023 to 15/10/2023). Therefore, as a solution, it is proposed to extend the series backwards using probability distributions, starting from the statistics of the original series. This extension allows initializing the anomaly detection models.

    \subsection{Data acquisition}
        Regarding data acquisition, video segments equivalent to 15 minutes are obtained. Due to the high amount of space required for storage, the resolution is reduced to 1280x720 pixels and the frame rate to 1 FPS.

    \subsection{Data processing}
    \label{subsec:dataprocessing}
        Video processing begins with the generation of a background model, based on Gaussian mixture models \cite{zivkovic_efficient_2006, zivkovic_improved_2004}, aimed at removing static objects from the image that could generate false positives. For each frame of the video, we update the background model and apply a pre-trained YOLOv8 model \cite{Jocher_Ultralytics_YOLO_2023} for object segmentation task. YOLO is an object detection model based on convolutional neural networks that processes entire images in a single stage to identify and locate objects. Specifically, the employed model provides us with information corresponding to both segmentation and object detection. We use the 'small' version of this model with a confidence threshold of $0.5$.
    
        To extract detections of dynamic objects, we compare the areas corresponding to the bounding boxes of all detections in the frame with those areas in the background model. These cutouts are in grayscale and undergo Contrast Limited Adaptive Histogram Equalization (CLAHE) processing \cite{puer_adaptive_nodate, pizer_contrast-limited_1990, books/el/94/Zuiderveld94}.
        For comparing the cutouts, we use the Structural Similarity Index (SSIM) \cite{zhou_wang_mean_2009, wang_image_2004}, which can take values between -1 and 1, with 1 indicating perfect similarity between images, and -1 indicating completely dissimilar images. In our case, we consider a heuristic threshold of $0.8$, considering detections valid when their SSIM value is below this threshold.

        The result of this process is stored in a '.csv' file for each video segment. This file will store information regarding the detection timestamp, the predicted class, prediction confidence, as well as the bounding box position and segmentation mask. No other video information is saved, avoiding any privacy concern with the data.

    \subsection{Generating the Time Series}
    \label{subsec:TSGeneration}
        From the result of the video processing, we generate two time series corresponding to the number of detections and the percentage of heatmap saturation, which represents the image occupancy within a given interval.
        
            \subsubsection{Detection Series}
                To generate the series corresponding to the number of detections, for each '.csv' file resulting from the video processing, we group the detections by time interval and obtain the maximum value within these groups. This value is then used in the time series to refer to that interval. Choosing the maximum over other statistics, such as mean or median, helps correct potential false negatives that may occur. In certain detection intervals, some people may not be detected due to either the precision of the YOLO model or the SSIM threshold; however, detections may occur throughout the interval.

\begin{figure}[h]
                    \centering
                    \includegraphics[scale=0.35]{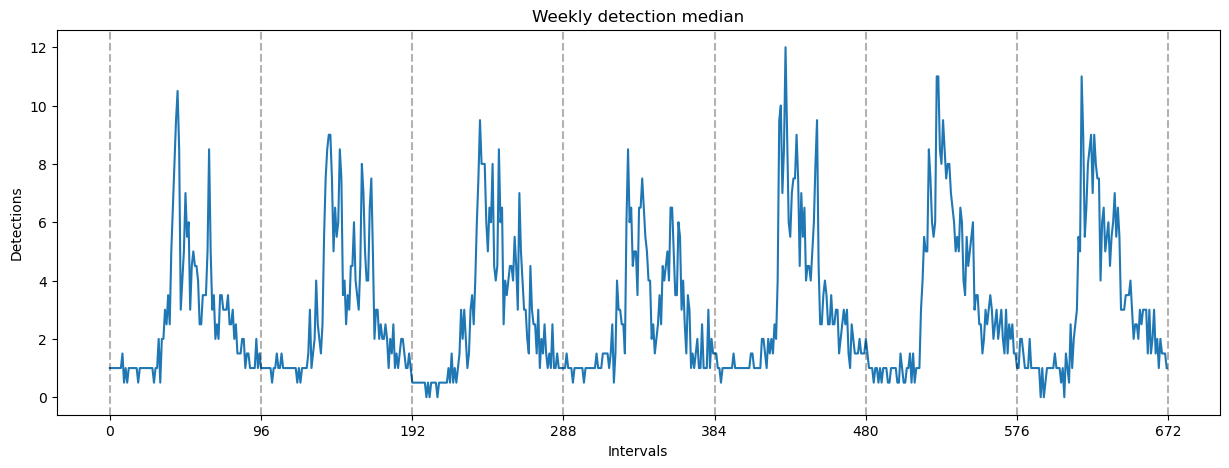}
                    \caption{Weekly distribution of median detection series}
                    \label{fig:median_detection}
                \end{figure}

\begin{figure}[h]
                    \centering
                    \includegraphics[scale=0.35]{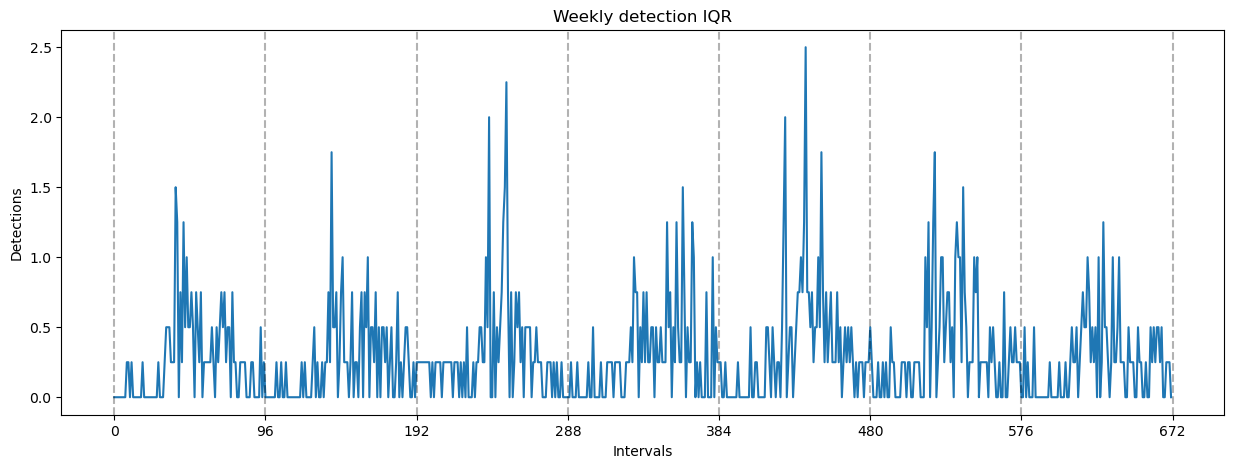}
                    \caption{Weekly distribution of IQR detection series}
                    \label{fig:IQR_detection}
                \end{figure}

                We take 50\% of the available data to extract statistics, aiming to increase the series. We group this partition by day of the week, hour, and minute (w, h, m), obtaining both the median (Figure \ref{fig:median_detection}) and the interquartile range (IQR) (Figure \ref{fig:IQR_detection}) for each of the groups.
    
                
                For modeling the artificial series, we employ a Gumbel distribution \cite{tocher_statistical_1955} (Equation \ref{eqn:Gumbel}). This distribution allows modeling the distribution of maximum or minimum values.
                
                \begin{equation}
                    \label{eqn:Gumbel}
                    Gumbel(x|\mu,\beta)=e^{-e^{-\frac{x-\mu}{\beta}}}
                \end{equation}
    
\begin{figure}[h]
                    \centering
                    \includegraphics[scale=0.35]{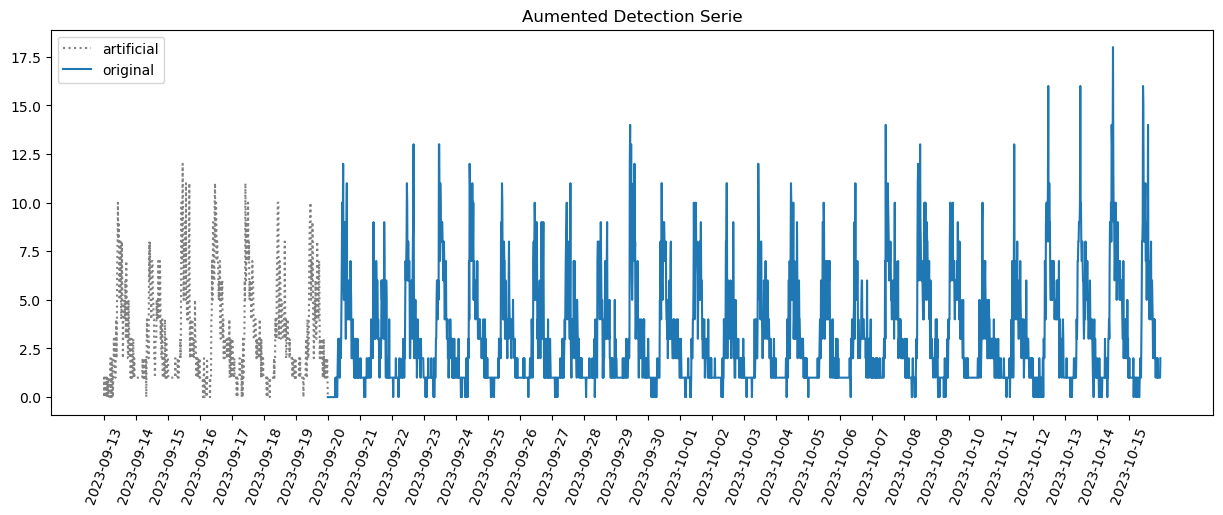}
                    \caption{Augmented detection series}
                    \label{fig:aumented_detection}
                \end{figure}

                We generate the values corresponding to 8 weeks with a frequency of 15 minutes, where $\mu$ is the median of the corresponding (w, h, m), and $\beta$ is the quartile deviation $(IQR/2)$ of the corresponding (w, h, m). (See Figure \ref{fig:aumented_detection})
    
            \subsubsection{Heatmap Saturation Percentage Series}
                For generating the series corresponding to the heatmap, for each '.csv' file resulting from the video processing, we use the segmentation masks of each detection.

                To obtain the grayscale heatmap, given a '.csv' file, we accumulate the segmentation masks of the different detections into a matrix initialized with zeros, of the same size as the original image. Once accumulated, we normalize the data so that, for a point in the map to saturate to white, it must have been occupied throughout the entire interval.
                
                Next, from these heatmaps, we generate the heatmap series. Each heatmap represents a point within the series, with this value being the sum of the values in the image. The theoretical maximum value for a heatmap is equal to \(|columns|*|rows|*255\), i.e., when it is completely saturated.
        
\begin{figure}[h]
                    \centering
                    \includegraphics[scale=0.35]{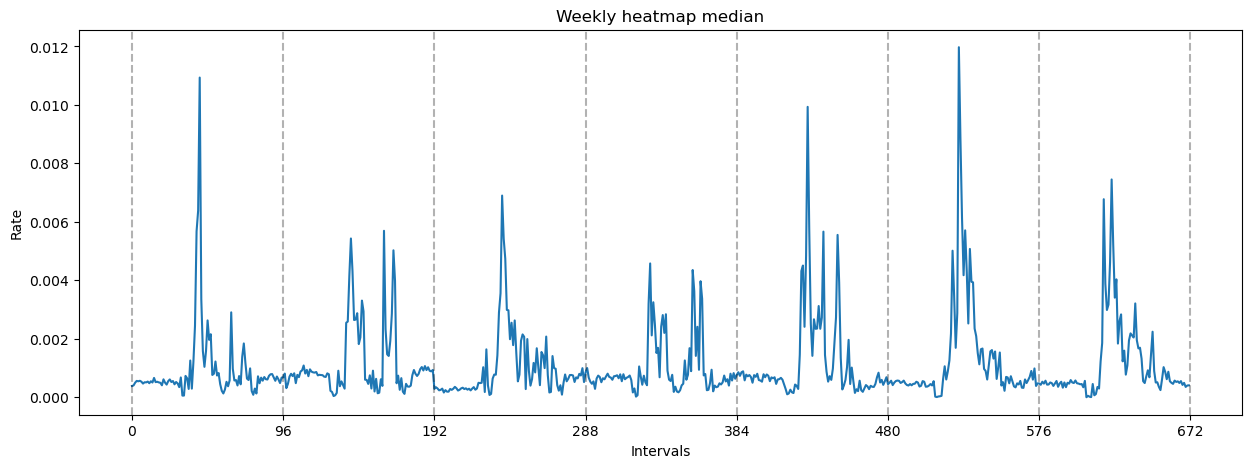}
                    \caption{Weekly distribution of the median of the heatmap series}
                    \label{fig:median_heatmap}
                \end{figure}

\begin{figure}[h]
                    \centering
                    \includegraphics[scale=0.35]{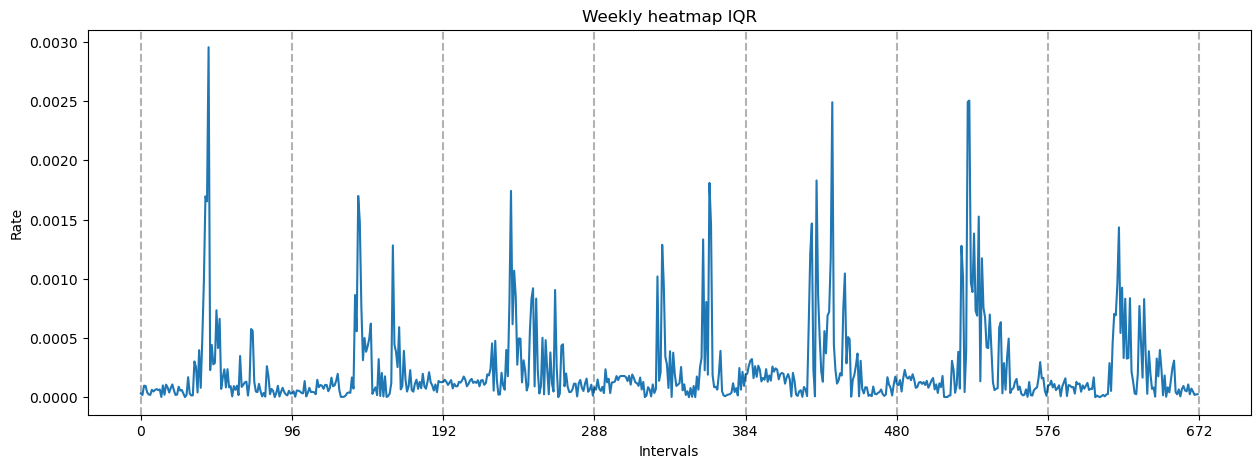}
                    \caption{Weekly distribution of the IQR of the heatmap series}
                    \label{fig:iqr_heatmap}
                \end{figure}

                We take 50\% of the available data to extract statistics, aiming to increase the series. We group this partition by day of the week, hour, and minute, obtaining both the median (Figure \ref{fig:median_heatmap}) and the interquartile range (Figure \ref{fig:iqr_heatmap}) for each.
    
        
                For modeling the artificial series, we use the Laplace distribution \cite{kotz_laplace_2001} (Equation \ref{eqn:LaPlace}). The choice of this distribution over others, such as the normal distribution, is due to its characteristics. Its sharper peak and faster decay allow us to control the amount of noise applied to the generated series.
    
                \begin{equation}
                    \label{eqn:LaPlace}
                    Laplace(x|\mu,\beta)=\frac{1}{2\beta}^{-\frac{|x-\mu|}{\beta}}
                \end{equation}
                
                Using a Laplace distribution, we generate the values corresponding to 8 weeks with a frequency of 15 minutes, where $\mu$ is the median of the corresponding (w, h, m), and $\beta$ is the quartile deviation of the corresponding (w, h, m).

\begin{figure}[h]
                    \centering
                    \includegraphics[scale=0.35]{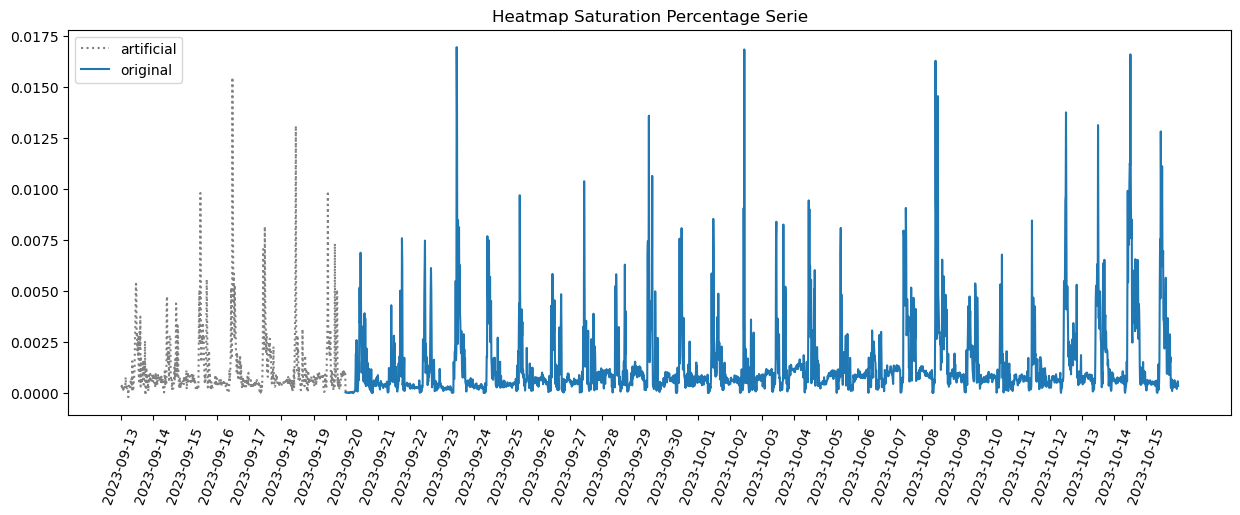}
                    \caption{Augmented heatmap saturation percentage series}
                    \label{fig:aumented_heatmap}
                \end{figure}

                Due to the high values of the series, we normalize it using the previously mentioned theoretical maximum value, thus representing the percentage of heatmap saturation (see Figure \ref{fig:aumented_heatmap}).

\section{Methodology}
\begin{figure}[h]
        \centering
        \includegraphics[width=\textwidth]{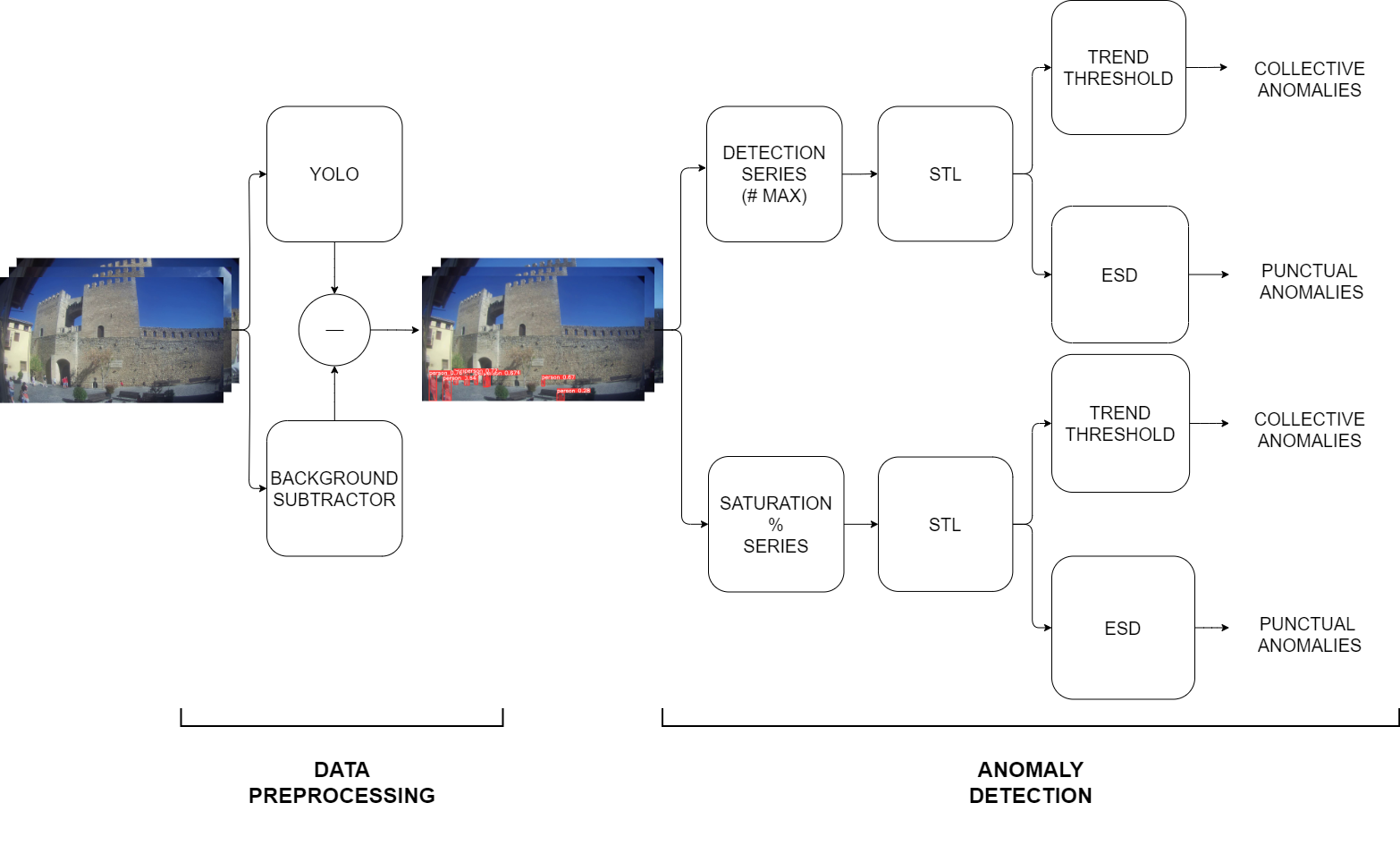}
        \caption{Diagram of the methodology employed}
        \label{fig:diagrama_metodologías}
    \end{figure}

    In Figure \ref{fig:diagrama_metodologías}, the methodology employed throughout the project can be observed.
    Part of this diagram has already been discussed in sections \ref{subsec:dataprocessing} and \ref{subsec:TSGeneration}, consisting of data preprocessing, while the remaining part is explained below. The central idea of this latter part is to apply anomaly detection techniques to the preprocessed data. Before applying these techniques, the series has been decomposed into its trend, seasonality, and residual components.

    \subsection{STL Decomposition}
        Through STL (Seasonal and Trend decomposition using Loess) \cite{cleveland90}, we aim to obtain the decomposition of the different series into trend, seasonality, and residual. This algorithm allows us to adjust parameters such as the periodicity of the series or its seasonality. Specifically, we use a periodicity of 1 day and a seasonality of 1 week.
    
    \subsection{Collective Anomalies - Trend Threshold}
        One of the types of anomalies we seek to identify are collective anomalies, i.e., those that individually do not represent an anomaly but do so when considered as a sequence.

        To achieve this, a threshold is set on the trend calculated in the series decomposition. For both series, this threshold $\delta$ is $\delta = \tilde{x} \pm \sigma$, where $\tilde{x}$ is the median of the original series and $\sigma$ is its standard deviation. We consider a value anomalous only if it exceeds the defined upper threshold.
            
    \subsection{Point Anomalies - SESD (Seasonal ESD)}
        Another type of anomaly we seek to identify are point anomalies, i.e., points that show a significant deviation from the rest of the data.

        For this, we employ the Seasonal Extreme Studentized Deviate (SESD) algorithm \cite{hochenbaum_automatic_2017}, which involves applying the ESD (Extreme Studentized Deviate) algorithm \cite{Rosner_1983} to the result obtained after performing the STL decomposition of the series, in our case, based on the previously calculated decomposition.

        When selecting point anomalies, we discard those that have been identified as collective anomalies in the previous section. Point anomalies are chosen in descending order of residual value, with the most relevant being those with a high residual value.

\section{Analysis of results}
    Next, we proceed to analyze the results obtained after applying the methods discussed earlier.
        
    \subsection{STL Decomposition}
        \subsubsection{Detection Series}
\begin{figure}[h]
                \centering
                \includegraphics[scale=0.35]{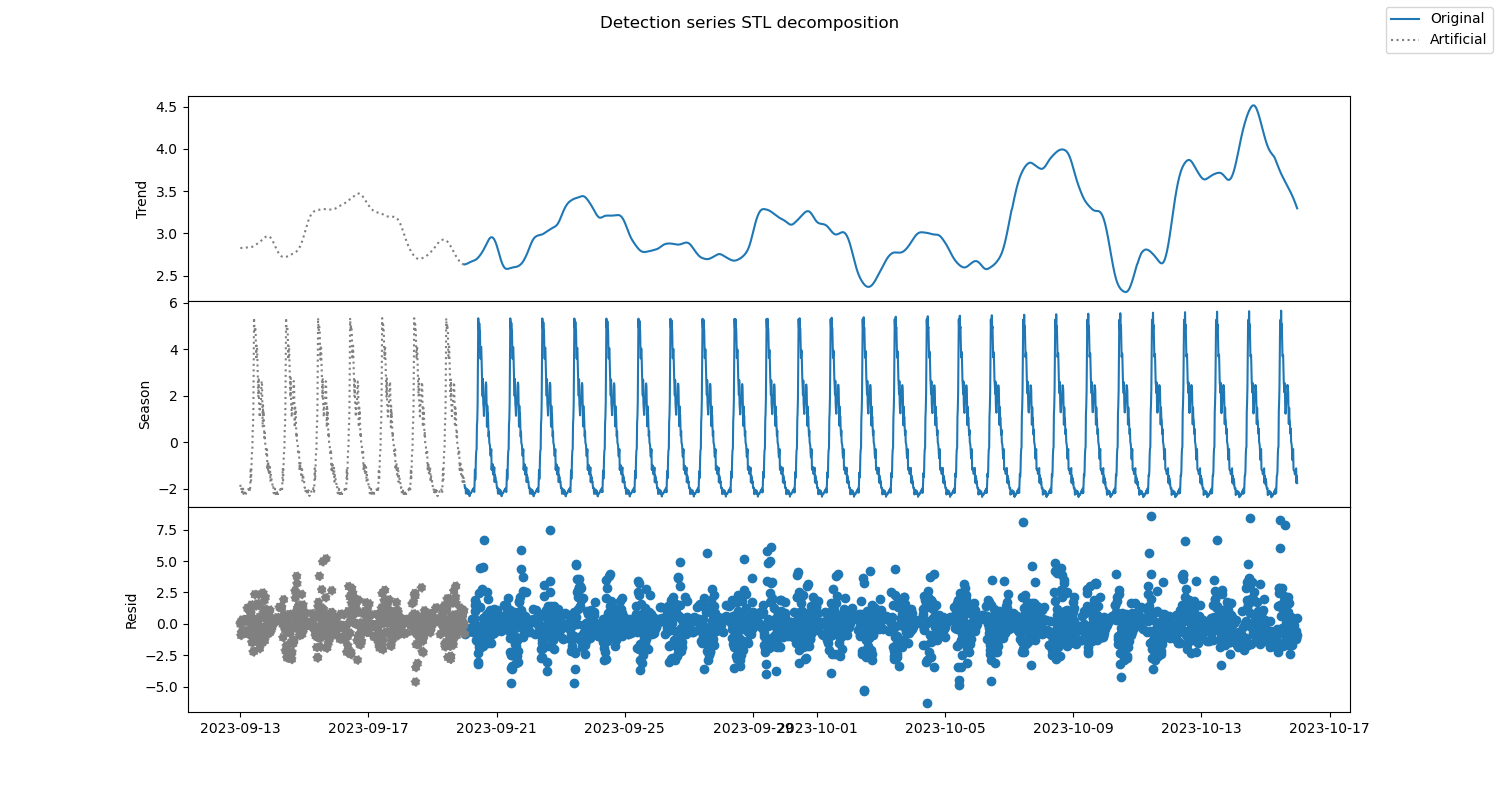}
                \caption{STL Decomposition of the Detection Series}
                \label{fig:STL_detection}
            \end{figure}

            Regarding the detection series, in Figure \ref{fig:STL_detection}, we can observe its STL decomposition. The seasonal component shows no variation over time, and the trend is flexible, with a noticeable change in trend present in the last week.
        
        \subsubsection{Heatmap Saturation Percentage Series}
\begin{figure}[h]
                \centering
                \includegraphics[scale=0.35]{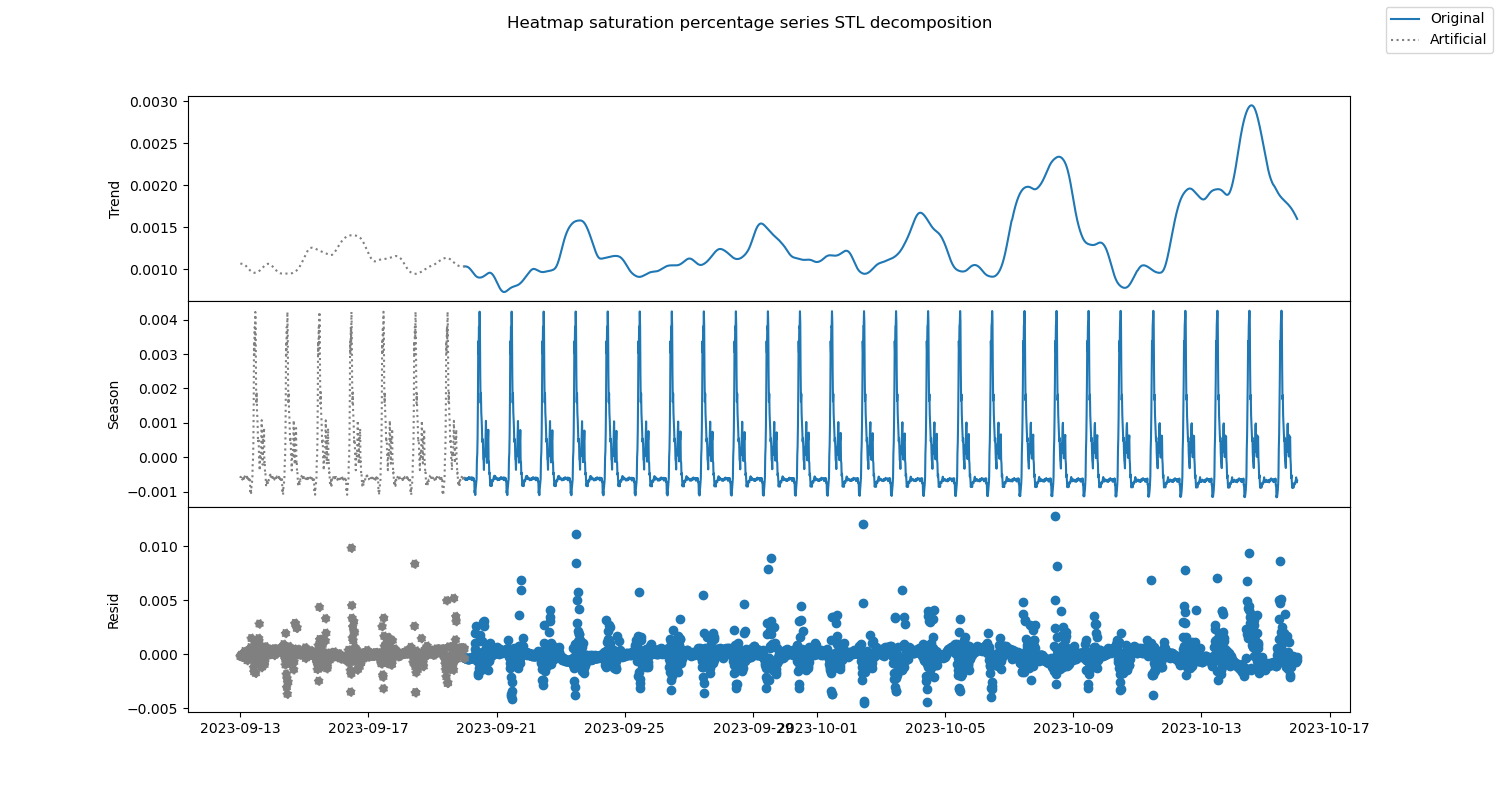}
                \caption{STL Decomposition of the Heatmap Saturation Percentage Series}
                \label{fig:STL_heatmap}
            \end{figure}

            Regarding the saturation percentage series of the heatmap, in Figure \ref{fig:STL_heatmap}, we can observe its STL decomposition. Similarly to the previous series, the seasonal component also shows no variation over time, and the trend is flexible, with the change in trend in the last week being more evident in this case.
            
    \subsection{Collective Anomalies - Trend Threshold}
        \subsubsection{Detection Series}
\begin{figure}[h]
                \centering
                \includegraphics[scale=0.35]{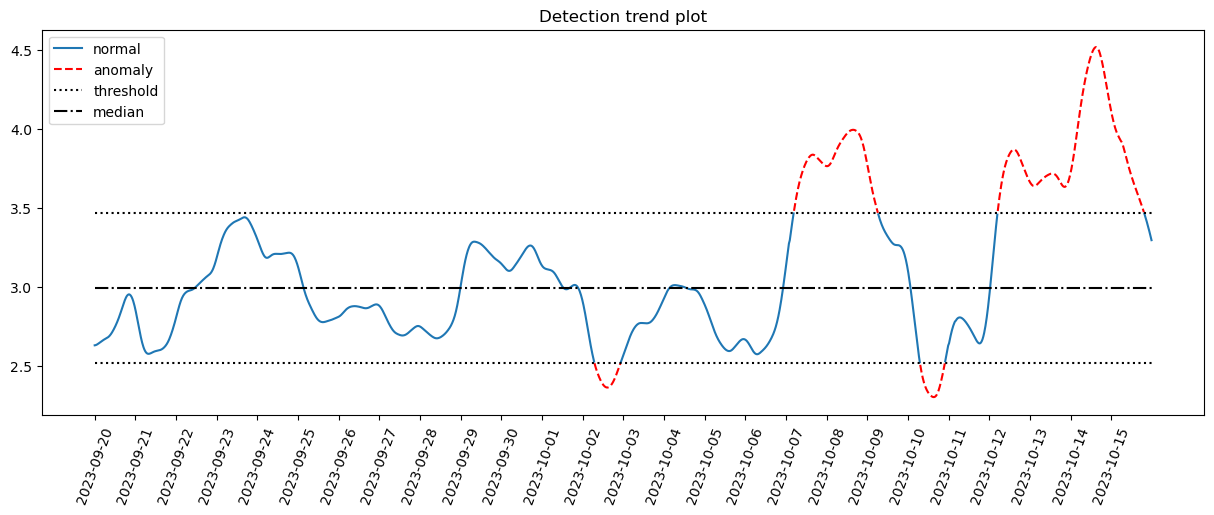}
                \caption{Trend Threshold in Detection Series}
                \label{fig:detection_trend_threshold}
            \end{figure}

            In Figure \ref{fig:detection_trend_threshold}, two sequences of anomalous points have been detected in the series corresponding to the number of detections. These sequences are between October 7th and 9th, and October 12th and 15th, representing festive periods in the \textit{Comunitat Valenciana} and nationally, respectively. Due to the tourist interest in the city of Morella combined with the holiday period, we can assume a higher influx of people, as we can observe in this figure.

        \subsubsection{Heatmap Saturation Percentage Series}
\begin{figure}[h]
                \centering
                \includegraphics[scale=0.35]{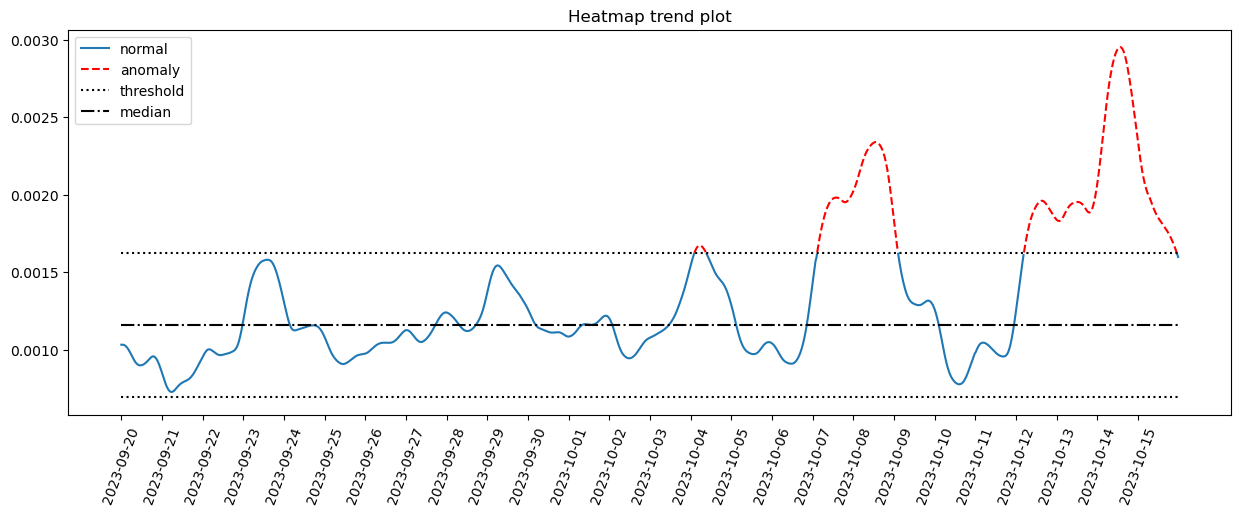}
                \caption{Trend Threshold in Heatmap Saturation Percenteage Series}
                \label{fig:heatmap_trend_threshold}
            \end{figure}

            In Figure \ref{fig:heatmap_trend_threshold}, corresponding to the saturation percentage series of the heatmap, the anomalous sequences detected in the detection series are found again, which coincide with the mentioned festive periods. However, in this series, another sequence corresponding to October 4th appears. This sequence can be considered anomalous due to the detection of false positives in the early hours of this day.
                
    \subsection{Point Anomalies - SESD (Seasonal ESD)}
        \subsubsection{Detection Series}
\begin{figure}[h]
                \centering
                \includegraphics[scale=0.35]{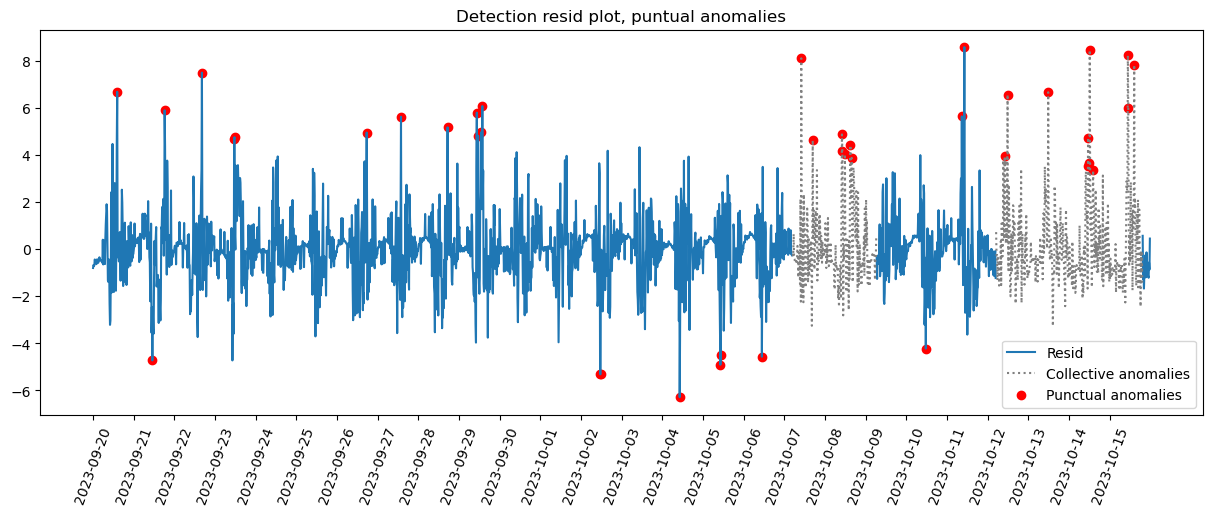}
                \caption{Plot of Detection Residual with Point Anomalies}
                \label{fig:detection_puntual}
            \end{figure}

            In Figure \ref{fig:detection_puntual}, we can observe the different point anomalies that have been detected for the detection series after applying the ESD algorithm to the STL decomposition of the series. Despite not considering the intervals considered as collective anomalies, the algorithm is able to detect anomalies in these intervals, reinforcing the validity of the methodology used.

\begin{figure}[h]
            \captionsetup[subfigure]{justification=centering}
                \begin{subfigure}{.35\textwidth}
                  \centering
                  \includegraphics[width=.8\linewidth]{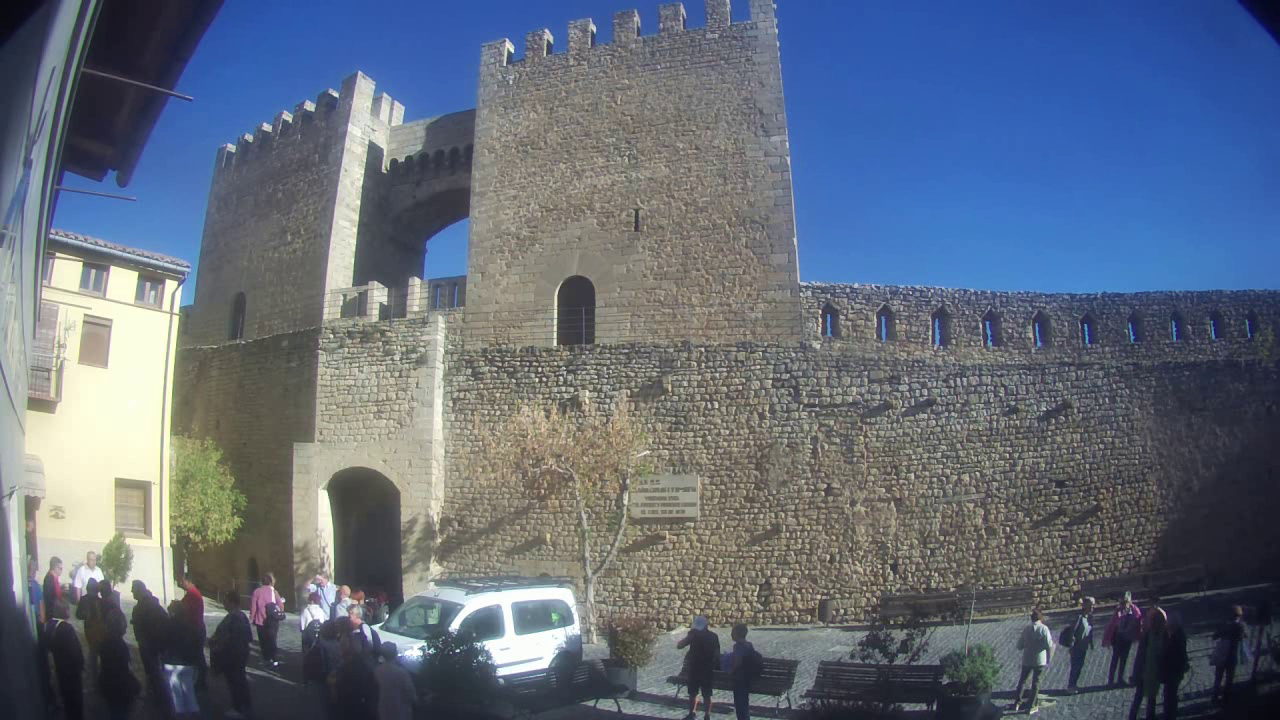}
                  \caption{'2023-10-11 10:15:00' \\ (Anomaly)}
                  \label{fig:detection_anomaly}
                \end{subfigure}%
                \begin{subfigure}{.35\textwidth}
                  \centering
                  \includegraphics[width=.8\linewidth]{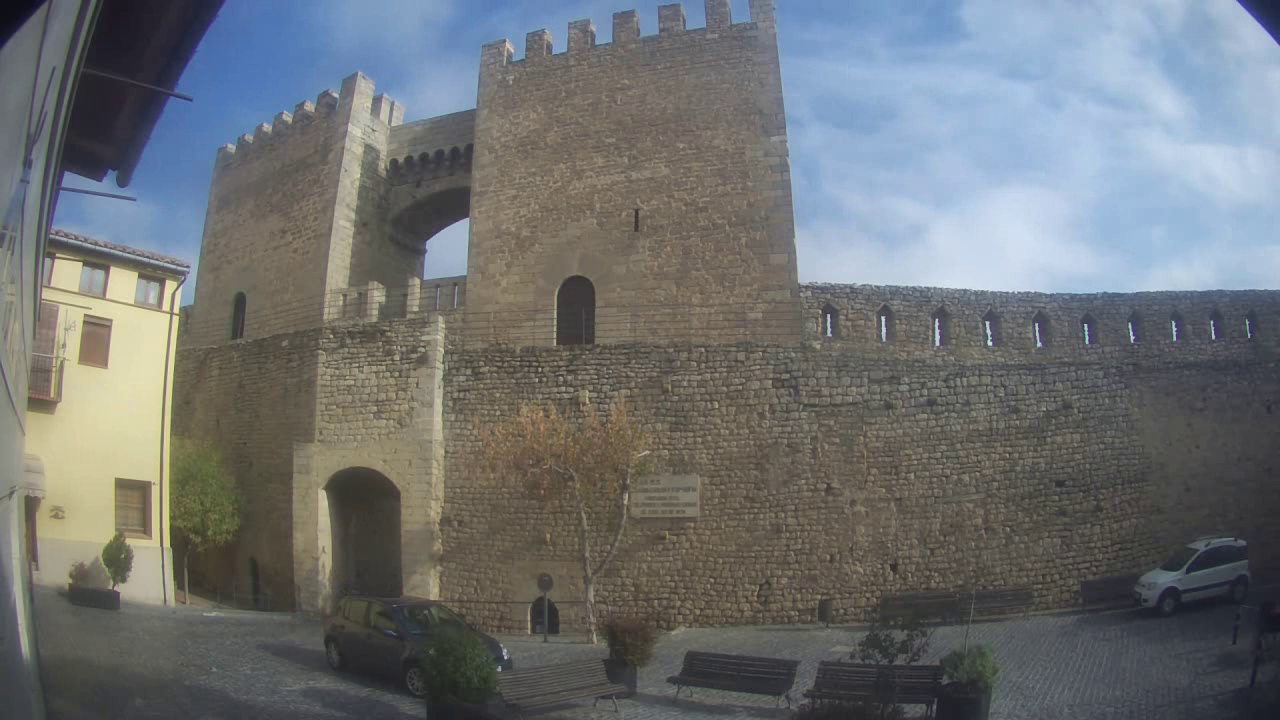}
                  \caption{'2023-10-04 10:15:00' \\ (Previous week)}
                  \label{fig:detection-1s}
                \end{subfigure}
                \begin{subfigure}{.35\textwidth}
                  \centering
                  \includegraphics[width=.8\linewidth]{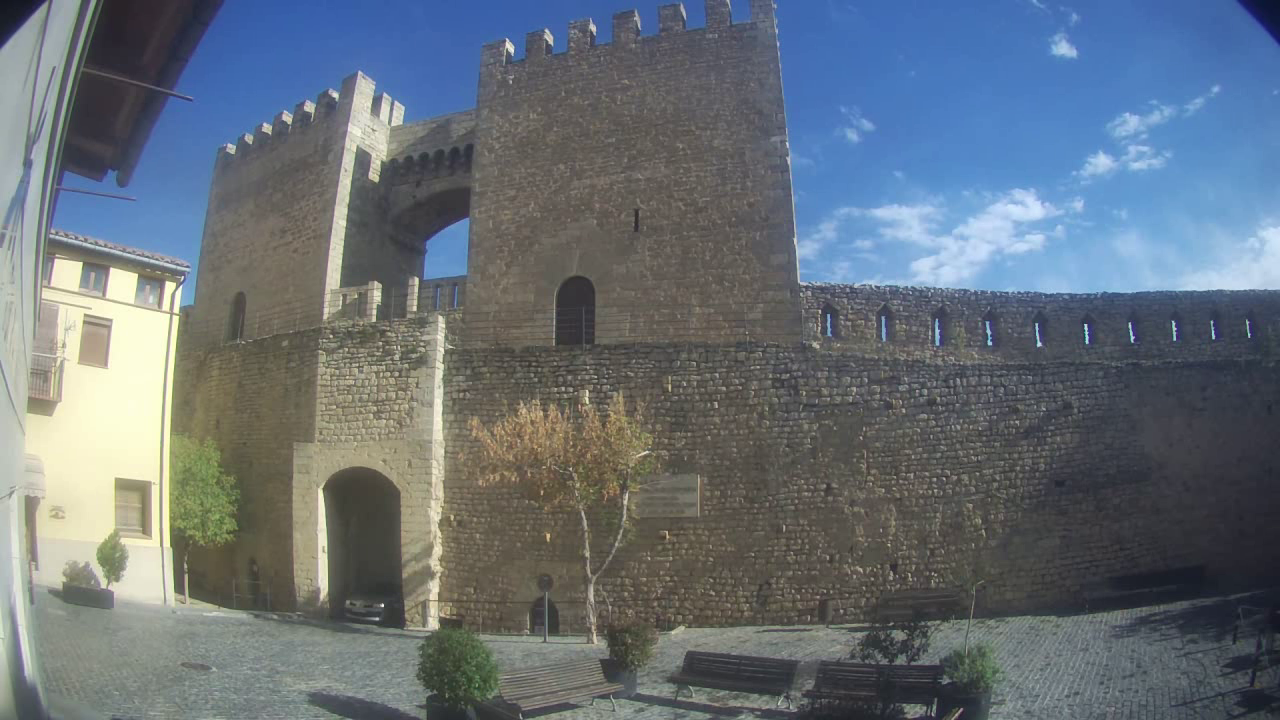}
                  \caption{'2023-09-20 10:15:00' \\ (3 weeks earlier)}
                  \label{fig:detection-3s}
                \end{subfigure}
                \caption{Justification of Anomalies in Detection Series}
                \label{fig:detection_screenshots}
            \end{figure}

            To validate the proposed point anomalies, the anomaly with the highest residue is verified with the available previous or subsequent instances. In Figure \ref{fig:detection_screenshots}, we can observe the capture corresponding to the anomalous value of '2023-10-11 10:15:00' compared to captures from 1 and 3 weeks earlier. In these, a clear change in the number of people present in the images can be observed. Checks have been carried out at more time instances and with different anomalies, but for the sake of article length, they are omitted.

        \subsubsection{Heatmap Saturation Percentage Series}
\begin{figure}[htbp]
                \centering
                \includegraphics[scale=0.35]{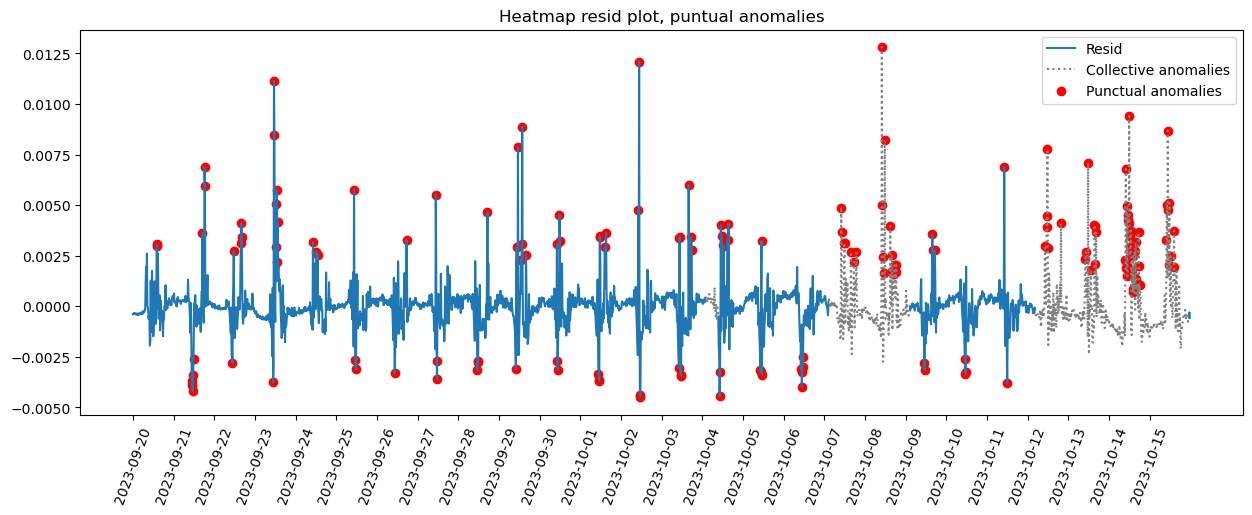}
                \caption{Plot of Heatmap Saturation Percentage Residual with Point Anomalies}
                \label{fig:heatmap_puntual}
            \end{figure}

            In Figure \ref{fig:heatmap_puntual}, we can observe the different point anomalies that have been detected for the saturation percentage series of the heatmap, after applying the ESD algorithm to the STL decomposition of the series. Despite not considering the intervals considered as collective anomalies, the algorithm is able to detect anomalies in these intervals, reinforcing the validity of the methodology used.

\begin{figure}[htbp]
            \captionsetup[subfigure]{justification=centering}
                \begin{subfigure}{.35\textwidth}
                  \centering
                  \includegraphics[width=.8\linewidth]{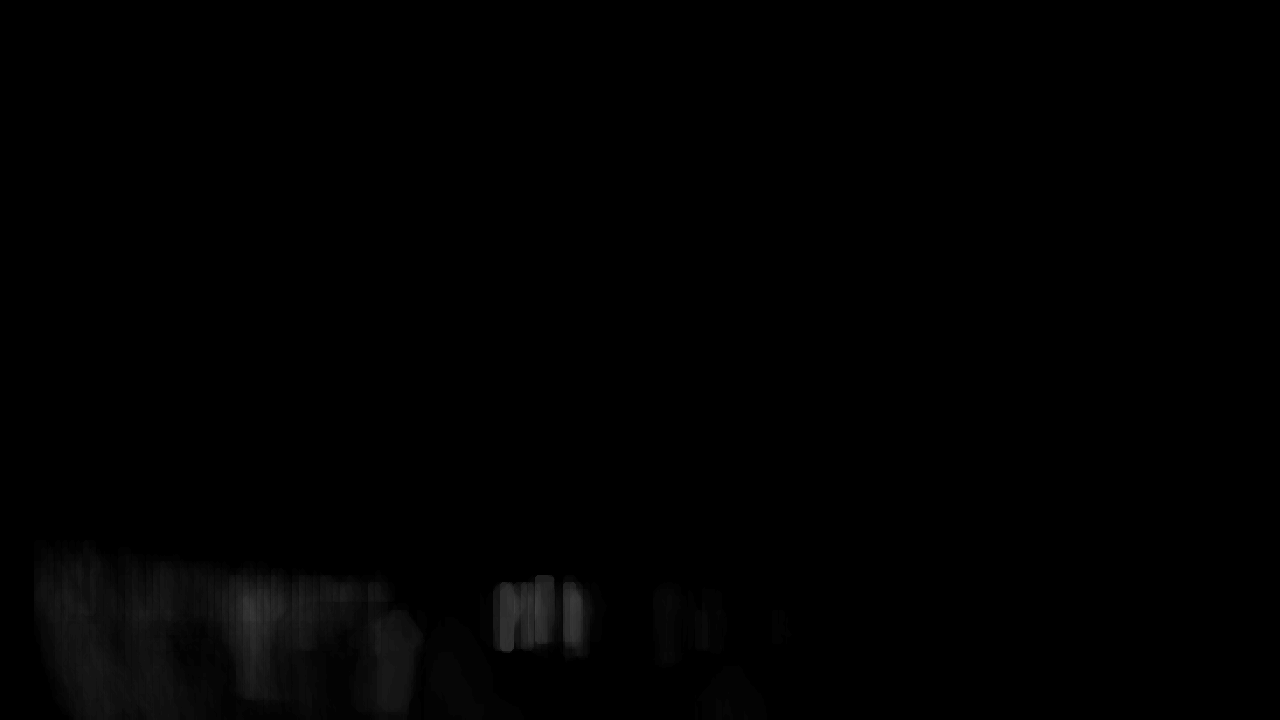}
                  \caption{'2023-10-01 10:45:00' \\ (Previous day) [0.001601]}
                  \label{fig:heatmap-1d}
                \end{subfigure}
                \begin{subfigure}{.35\textwidth}
                  \centering
                  \includegraphics[width=.8\linewidth]{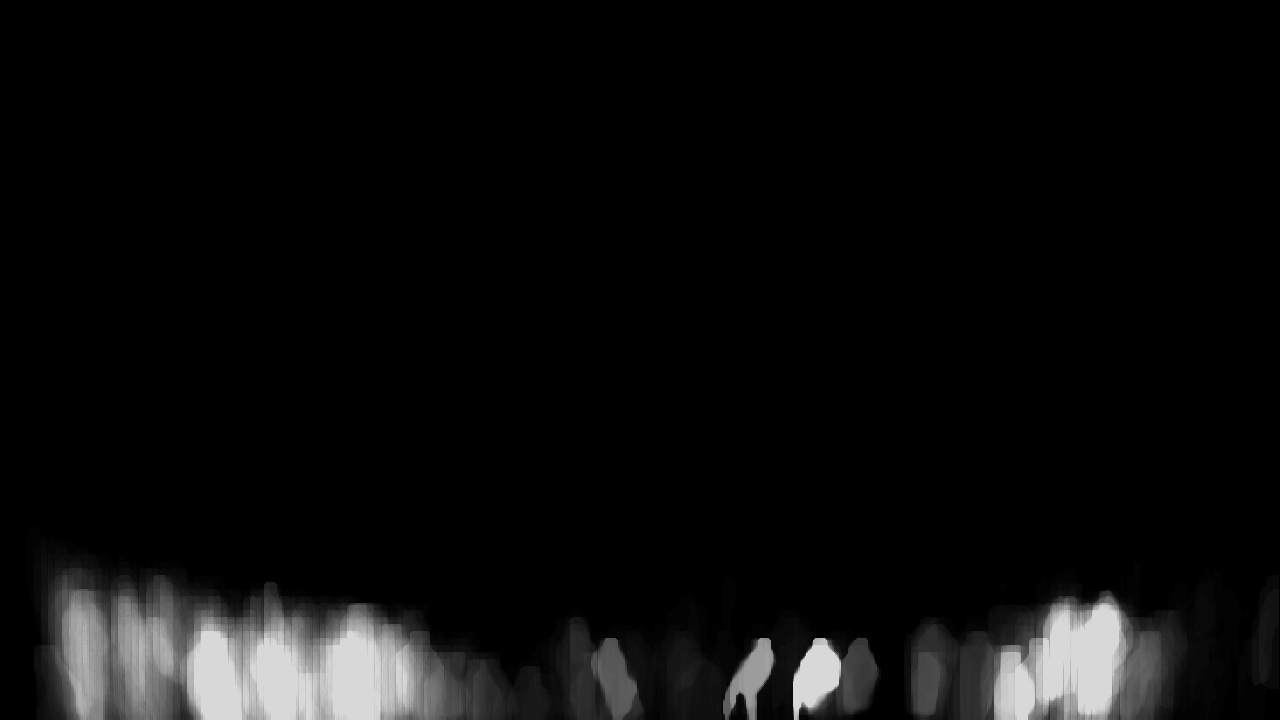}
                  \caption{'2023-10-02 10:45:00' \\ (Anomaly) [0.016843]}
                  \label{fig:heatmap_anomaly}
                \end{subfigure}
                \begin{subfigure}{.35\textwidth}
                  \centering
                  \includegraphics[width=.8\linewidth]{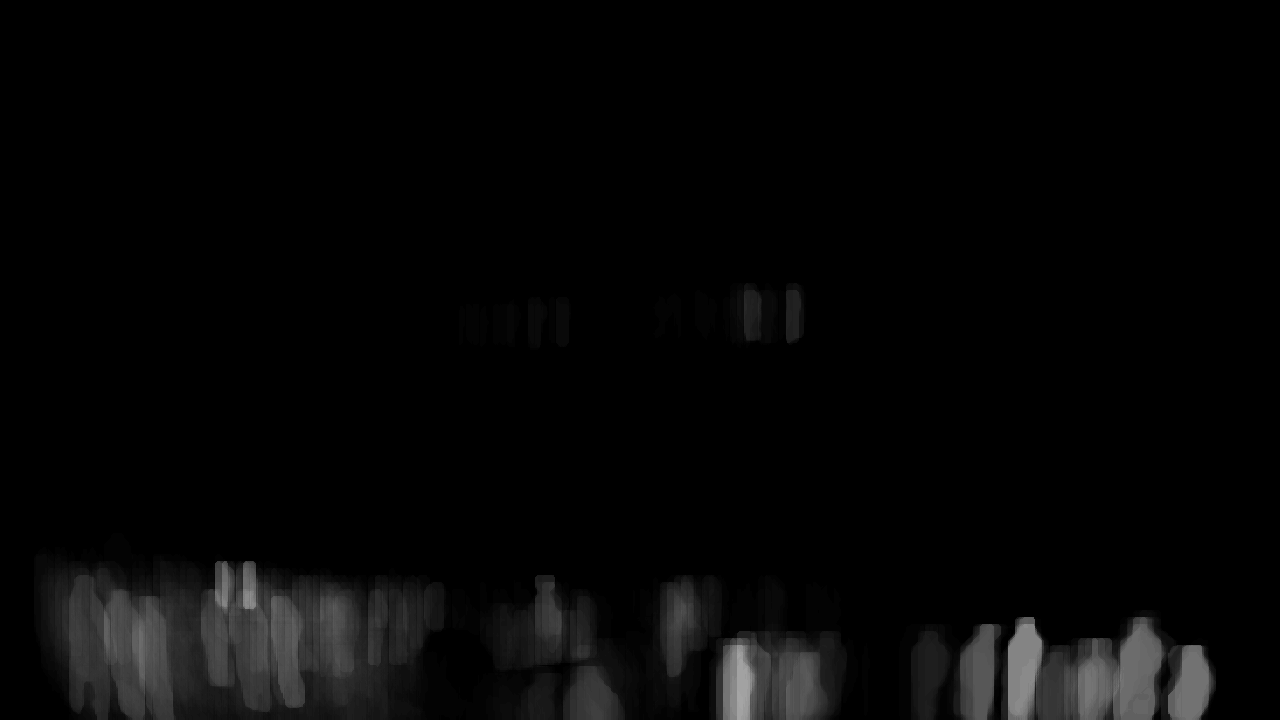}
                  \caption{'2023-10-03 10:45:00' \\ (Next day) [0.008394]}
                  \label{fig:heatmap+1d}
                \end{subfigure}
                \caption{Justification of Anomalies in Heatmap Series}
                \label{fig:heatmap_screenshots}
            \end{figure}

            To validate the proposed point anomalies, the anomaly with the highest residue is verified with the available previous or subsequent instances. In Figure \ref{fig:heatmap_screenshots}, we can observe the capture corresponding to the anomalous value of '2023-10-02 10:45:00' compared to captures from the previous and subsequent day. The difference in saturation is evident, obtaining a higher saturation value (0.016843) in the one considered anomalous. Checks have been carried out at more time instances and with different anomalies, but for the sake of article length, they are omitted.

\section{Discussion of Results}
    The results obtained reveal that the utilization of time series decomposition through STL has been effective in identifying seasonal patterns and flexible trends. Specifically, a significant change in trend is observed around the days 9 and 12 of October, coinciding with regional and national festivities respectively. This observation reinforces the usefulness of decomposition in capturing seasonal events and abrupt changes in the data. Additionally, by analyzing the residual of the decomposition, points that are distant from the rest are identified, indicative of the presence of anomalous values. The detection of collective anomalies through the application of a threshold to the trend confirms the presence of anomalous periods in both series, these being the festive periods mentioned previously. On the other hand, when using the SESD algorithm to detect point anomalies, the validity of the collective anomaly detection method is evidenced, as the intervals identified previously as anomalous also stand out as such. This analysis also suggests that other points detected as point anomalies represent significant changes in different time intervals, thus consolidating the effectiveness of the approach used for anomaly detection in the studied time series.

\section{Conclusions}
    In this article, the application of anomaly detection techniques in open data, specifically in video data from the \textit{Comunitat Valenciana}, has been explored. From these data, two time series are generated, which are expanded using probability distributions. Subsequently, anomaly detection techniques are applied to these expanded series. On one hand, collective anomalies are identified by establishing a threshold based on the statistics of each series and applying it to the trend derived from its decomposition. On the other hand, point anomalies are detected using the SESD algorithm applied to the same decomposition. To validate the obtained results, the most prominent point anomalies of each series are compared with different time intervals. These findings suggest the feasibility and usefulness of anomaly detection techniques in identifying anomalous patterns in temporal data, thus providing a valuable tool for open data analysis in video format in complex environments.

\section*{Acknowledgments}
    Grant PID2021-127946OB-I00 funded by MCIN/AEI/ 10.13039/501100011033 by “ERDF A way of making Europe”.

\printbibliography

@article{ding_mst-gat_2023,
	title = {{MST}-{GAT}: {A} multimodal spatial–temporal graph attention network for time series anomaly detection},
	volume = {89},
	issn = {15662535},
	shorttitle = {{MST}-{GAT}},
	url = {https://linkinghub.elsevier.com/retrieve/pii/S156625352200104X},
	doi = {10.1016/j.inffus.2022.08.011},
	language = {en},
	urldate = {2024-05-07},
	journal = {Information Fusion},
	author = {Ding, Chaoyue and Sun, Shiliang and Zhao, Jing},
	month = jan,
	year = {2023},
	pages = {527--536},
}

@InProceedings{pmlr-v95-guo18a,
  title = 	 {Multidimensional Time Series Anomaly Detection: A GRU-based Gaussian Mixture Variational Autoencoder Approach},
  author =       {Guo, Yifan and Liao, Weixian and Wang, Qianlong and Yu, Lixing and Ji, Tianxi and Li, Pan},
  booktitle = 	 {Proceedings of The 10th Asian Conference on Machine Learning},
  pages = 	 {97--112},
  year = 	 {2018},
  editor = 	 {Zhu, Jun and Takeuchi, Ichiro},
  volume = 	 {95},
  series = 	 {Proceedings of Machine Learning Research},
  month = 	 {14--16 Nov},
  publisher =    {PMLR},
  url = 	 {https://proceedings.mlr.press/v95/guo18a.html},
  %doi = 	 
}

@inproceedings{nedelkoski_anomaly_2019,
	address = {Milan, Italy},
	title = {Anomaly {Detection} from {System} {Tracing} {Data} {Using} {Multimodal} {Deep} {Learning}},
	copyright = {https://ieeexplore.ieee.org/Xplorehelp/downloads/license-information/IEEE.html},
	isbn = {978-1-72812-705-7},
	url = {https://ieeexplore.ieee.org/document/8814585/},
	doi = {10.1109/CLOUD.2019.00038},
	language = {en},
	urldate = {2024-05-07},
	booktitle = {2019 {IEEE} 12th {International} {Conference} on {Cloud} {Computing} ({CLOUD})},
	publisher = {IEEE},
	author = {Nedelkoski, Sasho and Cardoso, Jorge and Kao, Odej},
	month = jul,
	year = {2019},
	pages = {179--186},
}

@inbook{Shaukat_2021,
        title={{A} {Review} of {Time-Series} {Anomaly} {Detection} {Techniques}: {A} {Step} to {Future} {Perspectives}}, 
        ISBN={9783030731007}, 
        ISSN={2194-5365}, 
        url={http://dx.doi.org/10.1007/978-3-030-73100-7_60}, 
        DOI={10.1007/978-3-030-73100-7_60}, 
        booktitle={Advances in Information and Communication}, 
        publisher={Springer International Publishing}, 
        author={Shaukat, Kamran and Alam, Talha Mahboob and Luo, Suhuai and Shabbir, Shakir and Hameed, Ibrahim A. and Li, Jiaming and Abbas, Syed Konain and Javed, Umair}, 
        year={2021}, 
        pages={865–877}
}

@inproceedings{ren_time-series_2019,
	address = {Anchorage AK USA},
	title = {Time-{Series} {Anomaly} {Detection} {Service} at {Microsoft}},
	isbn = {978-1-4503-6201-6},
	url = {https://dl.acm.org/doi/10.1145/3292500.3330680},
	doi = {10.1145/3292500.3330680},
	language = {en},
	urldate = {2024-05-07},
	booktitle = {Proceedings of the 25th {ACM} {SIGKDD} {International} {Conference} on {Knowledge} {Discovery} \& {Data} {Mining}},
	publisher = {ACM},
	author = {Ren, Hansheng and Xu, Bixiong and Wang, Yujing and Yi, Chao and Huang, Congrui and Kou, Xiaoyu and Xing, Tony and Yang, Mao and Tong, Jie and Zhang, Qi},
	month = jul,
	year = {2019},
	pages = {3009--3017},
}

@inproceedings{laptev_generic_2015,
	address = {Sydney NSW Australia},
	title = {Generic and {Scalable} {Framework} for {Automated} {Time}-series {Anomaly} {Detection}},
	isbn = {978-1-4503-3664-2},
	url = {https://dl.acm.org/doi/10.1145/2783258.2788611},
	doi = {10.1145/2783258.2788611},
	language = {en},
	urldate = {2024-05-07},
	booktitle = {Proceedings of the 21th {ACM} {SIGKDD} {International} {Conference} on {Knowledge} {Discovery} and {Data} {Mining}},
	publisher = {ACM},
	author = {Laptev, Nikolay and Amizadeh, Saeed and Flint, Ian},
	month = aug,
	year = {2015},
	pages = {1939--1947},
}

@inproceedings{geiger_tadgan_2020,
	address = {Atlanta, GA, USA},
	title = {{TadGAN}: {Time} {Series} {Anomaly} {Detection} {Using} {Generative} {Adversarial} {Networks}},
	copyright = {https://ieeexplore.ieee.org/Xplorehelp/downloads/license-information/IEEE.html},
	isbn = {978-1-72816-251-5},
	shorttitle = {{TadGAN}},
	url = {https://ieeexplore.ieee.org/document/9378139/},
	doi = {10.1109/BigData50022.2020.9378139},
	language = {en},
	urldate = {2024-05-07},
	booktitle = {2020 {IEEE} {International} {Conference} on {Big} {Data} ({Big} {Data})},
	publisher = {IEEE},
	author = {Geiger, Alexander and Liu, Dongyu and Alnegheimish, Sarah and Cuesta-Infante, Alfredo and Veeramachaneni, Kalyan},
	month = dec,
	year = {2020},
	pages = {33--43},
}

@misc{xu_anomaly_2022,
	title = {Anomaly {Transformer}: {Time} {Series} {Anomaly} {Detection} with {Association} {Discrepancy}},
	shorttitle = {Anomaly {Transformer}},
	url = {http://arxiv.org/abs/2110.02642},
	language = {en},
	urldate = {2024-05-07},
	publisher = {arXiv},
	author = {Xu, Jiehui and Wu, Haixu and Wang, Jianmin and Long, Mingsheng},
	month = jun,
	year = {2022},
	note = {arXiv:2110.02642 [cs]},
    doi = {10.48550/arXiv.2110.02642},
}

@article{zivkovic_efficient_2006,
	title = {Efficient adaptive density estimation per image pixel for the task of background subtraction},
	volume = {27},
	copyright = {https://www.elsevier.com/tdm/userlicense/1.0/},
	issn = {01678655},
	url = {https://linkinghub.elsevier.com/retrieve/pii/S0167865505003521},
	doi = {10.1016/j.patrec.2005.11.005},
	language = {en},
	number = {7},
	urldate = {2024-05-07},
	journal = {Pattern Recognition Letters},
	author = {Zivkovic, Zoran and Van Der Heijden, Ferdinand},
	month = may,
	year = {2006},
	pages = {773--780},
}

@inproceedings{zivkovic_improved_2004,
	address = {Cambridge, UK},
	title = {Improved adaptive {Gaussian} mixture model for background subtraction},
	isbn = {978-0-7695-2128-2},
	url = {http://ieeexplore.ieee.org/document/1333992/},
	doi = {10.1109/ICPR.2004.1333992},
	language = {en},
	urldate = {2024-05-07},
	booktitle = {Proceedings of the 17th {International} {Conference} on {Pattern} {Recognition}, 2004. {ICPR} 2004.},
	publisher = {IEEE},
	author = {Zivkovic, Z.},
	year = {2004},
	pages = {28--31 Vol.2},
}

@misc{Jocher_Ultralytics_YOLO_2023,
        author = {Jocher, Glenn and Chaurasia, Ayush and Qiu, Jing},
        license = {AGPL-3.0},
        month = jan,
        title = {{Ultralytics YOLO}},
        url = {https://github.com/ultralytics/ultralytics},
        version = {8.0.0},
        year = {2023}
}

@article{puer_adaptive_nodate,
title = {Adaptive histogram equalization and its variations},
journal = {Computer Vision, Graphics, and Image Processing},
volume = {39},
number = {3},
pages = {355-368},
year = {1987},
issn = {0734-189X},
doi = {https://doi.org/10.1016/S0734-189X(87)80186-X},
url = {https://www.sciencedirect.com/science/article/pii/S0734189X8780186X},
author = {Stephen M. Pizer and E. Philip Amburn and John D. Austin and Robert Cromartie and Ari Geselowitz and Trey Greer and Bart {ter Haar Romeny} and John B. Zimmerman and Karel Zuiderveld},
}

@inproceedings{pizer_contrast-limited_1990,
	address = {Atlanta, GA, USA},
	title = {Contrast-limited adaptive histogram equalization: speed and effectiveness},
	isbn = {978-0-8186-2039-3},
	shorttitle = {Contrast-limited adaptive histogram equalization},
	url = {http://ieeexplore.ieee.org/document/109340/},
	doi = {10.1109/VBC.1990.109340},
	language = {en},
	urldate = {2024-05-07},
	booktitle = {[1990] {Proceedings} of the {First} {Conference} on {Visualization} in {Biomedical} {Computing}},
	publisher = {IEEE Comput. Soc. Press},
	author = {Pizer, S.M. and Johnston, R.E. and Ericksen, J.P. and Yankaskas, B.C. and Muller, K.E.},
	year = {1990},
	pages = {337--345},
}

@incollection{books/el/94/Zuiderveld94,
    author = {Zuiderveld, Karel J.},
    booktitle = {Graphics Gems},
    editor = {Heckbert, Paul S.},
    ee = {https://doi.org/10.1016/b978-0-12-336156-1.50061-6},
    isbn = {9780123361561},
    pages = {474-485},
    publisher = {Elsevier},
    title = {{Contrast} {Limited} {Adaptive} {Histogram} {Equalization}.},
    year = 1994
}

@article{zhou_wang_mean_2009,
	title = {Mean squared error: {Love} it or leave it? {A} new look at {Signal} {Fidelity} {Measures}},
	volume = {26},
	copyright = {https://ieeexplore.ieee.org/Xplorehelp/downloads/license-information/IEEE.html},
	issn = {1053-5888},
	shorttitle = {Mean squared error},
	url = {http://ieeexplore.ieee.org/document/4775883/},
	doi = {10.1109/MSP.2008.930649},
	language = {en},
	number = {1},
	urldate = {2024-05-07},
	journal = {IEEE Signal Processing Magazine},
	author = {{Zhou Wang} and Bovik, A.C.},
	month = jan,
	year = {2009},
	pages = {98--117},
}

@article{wang_image_2004,
	title = {Image {Quality} {Assessment}: {From} {Error} {Visibility} to {Structural} {Similarity}},
	volume = {13},
	copyright = {https://ieeexplore.ieee.org/Xplorehelp/downloads/license-information/IEEE.html},
	issn = {1057-7149},
	shorttitle = {Image {Quality} {Assessment}},
	url = {http://ieeexplore.ieee.org/document/1284395/},
	doi = {10.1109/TIP.2003.819861},
	language = {en},
	number = {4},
	urldate = {2024-05-07},
	journal = {IEEE Transactions on Image Processing},
	author = {Wang, Z. and Bovik, A.C. and Sheikh, H.R. and Simoncelli, E.P.},
	month = apr,
	year = {2004},
	pages = {600--612},
}

@article{tocher_statistical_1955,
	title = {Statistical {Theory} of {Extreme} {Values} and {Some} {Practical} {Applications}.},
	volume = {118},
	issn = {00359238},
	url = {https://www.jstor.org/stable/2342529?origin=crossref},
	doi = {10.2307/2342529},
	language = {en},
	number = {1},
	urldate = {2024-05-07},
	journal = {Journal of the Royal Statistical Society. Series A (General)},
	author = {Tocher, K. D. and Gumbel, E. J.},
	year = {1955},
	pages = {106},
}

@book{kotz_laplace_2001,
	address = {Boston, MA},
	title = {The {Laplace} {Distribution} and {Generalizations}},
	copyright = {http://www.springer.com/tdm},
	isbn = {978-1-4612-6646-4 978-1-4612-0173-1},
	url = {http://link.springer.com/10.1007/978-1-4612-0173-1},
	language = {en},
	urldate = {2024-05-07},
	publisher = {Birkhäuser Boston},
	author = {Kotz, Samuel and Kozubowski, Tomaz J. and Podgórski, Krzysztof},
	year = {2001},
	doi = {10.1007/978-1-4612-0173-1},
}

@article{cleveland90,
        author = {Cleveland, Robert B. and Cleveland, William S. and McRae, Jean E. and Terpenning, Irma},
        journal = {Journal of Official Statistics},
        pages = {3--73},
        title = {{STL}: {A} {Seasonal-Trend} {Decomposition} {Procedure} {Based} on {Loess}},
        volume = 6,
        year = 1990
}

@misc{hochenbaum_automatic_2017,
	title = {Automatic {Anomaly} {Detection} in the {Cloud} {Via} {Statistical} {Learning}},
	url = {http://arxiv.org/abs/1704.07706},
	language = {en},
	urldate = {2024-05-07},
	publisher = {arXiv},
	author = {Hochenbaum, Jordan and Vallis, Owen S. and Kejariwal, Arun},
	month = apr,
	year = {2017},
	note = {arXiv:1704.07706 [cs]},
}

@article{Rosner_1983,
        title={{Percentage} {Points} for a {Generalized} {ESD} {Many-Outlier} {Procedure}},
        volume={25},
        ISSN={0040-1706},
        url={http://dx.doi.org/10.2307/1268549},
        DOI={10.2307/1268549},
        number={2},
        journal={Technometrics},
        publisher={JSTOR},
        author={Rosner, Bernard},
        year={1983},
        month=may,
        pages={165}
}

@article{nazir_suspicious_2023,
	title = {Suspicious {Behavior} {Detection} with {Temporal} {Feature} {Extraction} and {Time}-{Series} {Classification} for {Shoplifting} {Crime} {Prevention}},
	volume = {23},
	copyright = {https://creativecommons.org/licenses/by/4.0/},
	issn = {1424-8220},
	url = {https://www.mdpi.com/1424-8220/23/13/5811},
	doi = {10.3390/s23135811},
	language = {en},
	number = {13},
	urldate = {2024-05-13},
	journal = {Sensors},
	author = {Nazir, Amril and Mitra, Rohan and Sulieman, Hana and Kamalov, Firuz},
	month = jun,
	year = {2023},
	pages = {5811},
}

@inproceedings{xue_real-time_2020,
	address = {Boston, MA, USA},
	title = {Real-{Time} {Anomaly} {Detection} and {Feature} {Analysis} {Based} on {Time} {Series} for {Surveillance} {Video}},
	copyright = {https://ieeexplore.ieee.org/Xplorehelp/downloads/license-information/IEEE.html},
	isbn = {978-1-72819-523-0},
	url = {https://ieeexplore.ieee.org/document/9426191/},
	doi = {10.1109/UV50937.2020.9426191},
	language = {en},
	urldate = {2024-05-13},
	booktitle = {2020 5th {International} {Conference} on {Universal} {Village} ({UV})},
	publisher = {IEEE},
	author = {Xue, Ruoyu and Chen, Jingyuan and Fang, Yajun},
	month = oct,
	year = {2020},
	pages = {1--7},
}

\end{document}